# Data Mixing for Large Language Models Pretraining: A Survey and Outlook

**Zhuo Chen[1,2], Yuxuan Miao[1], Supryadi[2], Deyi Xiong[2*]**

[1]The International Joint Institute of Tianjin University, Tianjin University, Tianjin 300072, China

[2]TJUNLP Lab, School of Computer Science and Technology, Tianjin University, Tianjin 300350, China

---

## ABSTRACT

Large language models (LLMs) rely on pretraining on massive and highly heterogeneous corpora, where the composition of training data has a decisive impact on training efficiency and downstream generalization under realistic compute and data budget constraints. Unlike samplelevel data selection, data mixing optimizes domain-level sampling weights to allocate limited budgets more effectively. In recent years, a growing body of work has proposed principled data mixing methods for LLMs pretraining; however, the literature remains fragmented and lacks a dedicated, systematic survey. This paper provides a comprehensive review of data mixing for LLMs pretraining. We first formalize data mixture optimization as a bilevel problem on the probability simplex and clarify the role of data mixing in the pretraining pipeline, and briefly explain how existing methods make this bilevel formulation tractable in practice. We then introduce a fine-grained taxonomy that organizes existing methods along two main dimensions: static versus dynamic mixing. Static mixing is further categorized into rule-based and learning-based methods, while dynamic mixing is further grouped into adaptive and externally guided families. For each class of methods, we summarize representative approaches and analyze their characteristics, strengths, and limitations from a performance-cost trade-off perspective. Building on this analysis, we highlight key challenges that cut across methods, including limited transferability across data domains, optimization objectives, models, and validation sets, as well as unstandardized evaluation protocols and benchmarks, as well as the inherent tension between performance gains and cost control in learning-based methods. Finally, we outline several exploratory directions, including finer-grained domain partitioning and inverse data mixing, as well as pipeline-aware designs, aiming to provide conceptual and methodological insights for future research.

---

[*]Corresponding author: Deyi Xiong (E-mail: dyxiong@tju.edu.cn; ORCID: 0000-0002-2353-5038).

## 1. INTRODUCTION

In recent years, large language models (LLMs) have achieved significant advances in natural language processing and, by virtue of strong language understanding and generation capabilities, have substantially propelled the development of artificial intelligence [1-5]. The core of LLMs development lies in Transformer [6] -based pretraining: during this stage, the model learns general-purpose language representations from massive corpora, laying the foundation for basic capabilities and generality. However, the accompanying training costs are high, including both computational expenses and the costs of data acquisition and processing. Therefore, under practical resource constraints, conducting efficient pretraining within a limited budget has become a pressing problem.

Existing research shows that the quality and composition of pretraining data significantly affect pretraining efficiency; high-quality, diverse, and distributionally balanced data can substantially enhance model generalization and downstream task performance [7-10].

Accordingly, from a data-centric perspective, efficiency can be improved along two complementary directions. The first is data selection: typically operating at the sample level or finer granularity, it filters data according to explicit quality criteria or scoring mechanisms to obtain a high-quality subset whose performance is comparable to that of the original dataset. The second is data mixing: typically operating at the domain level (with domains often defined by data sources), it strategically allocates sampling weights across domains under a given budget to make fuller use of existing data and improve overall performance. In practice, data selection and data mixing are often used together to further enhance pretraining efficiency. Beyond pretraining efficiency in the narrow sense, the choice of data mixture can also indirectly shape later stages such as alignment, safety tuning, and multimodal extension, because it determines which domains, styles, and interaction patterns the model is primarily exposed to during pretraining. For instance, mixtures that overemphasize narrow or biased sources may complicate subsequent safety alignment, whereas mixtures that better cover conversational, instruction-following, and multimodal-related data can provide a more robust foundation for later-stage objectives.

This paper focuses on data mixing, an emerging topic that has achieved breakthrough progress in recent years. Unlike earlier practices that mainly relied on trial-and-error or heuristic rules, recent years has seen the emergence of a series of principled methods that provide a more systematic framework for modeling and solving data mixture optimization problems. To the best of our knowledge, there remains a lack of specialized and comprehensive surveys on this topic. To fill this gap, we conduct a systematic review based on existing research: Section 2 presents a clear formulation of the data mixture optimization problem; Section 3 proposes a more comprehensive and finer-grained taxonomy that differs from prior work and, on this basis, offers a structured introduction to existing methods; Section 4 summarizes current limitations and challenges and provides a preliminary outlook on future directions. This paper aims to provide a clear problem characterization and a methodological landscape, and to further offer reusable conceptual and analytical frameworks that support systematic comparison and extension in subsequent work.

## 2. FORMULATION

To facilitate understanding and align with prior studies, we present a unified formulation of *data mixture optimization* in the context of LLMs pretraining. Let the pretraining corpus $\mathcal{D}$ partitioned into $n$ domains $\mathcal{D}_1, \ldots, \mathcal{D}_n$. We define a data mixture as a weight vector $\mathbf{k} = (k_1, \ldots, k_n) \in \Delta_n$ on the probability simplex:

$$\Delta_n := \{\mathbf{k} \in \mathbb{R}^n \mid \sum_{i=1}^{n} k_i = 1, \ k_i \geq 0 \ \forall_i\} \tag{1}$$

The components of $\mathbf{k}$ specify the sampling probability of each domain during training. The induced training distribution is given by the convex combination

$$\mathbb{P}_{\mathbf{k}} = \sum_{i=1}^{n} k_i \mathbb{P}_i \tag{2}$$

where $\mathbb{P}_i$ denotes the empirical (finite-sample) distribution of domain $\mathcal{D}_i$.

Different choices of $\mathbf{k}$ can substantially affect the final performance of the pretrained model. The goal of data mixture optimization is to identify a mixture that maximizes model performance; in practice, performance is commonly measured via an evaluation loss on a held-out set. Let $\theta(\mathbf{k})$ denote the parameters obtained by training under mixture $\mathbf{k}$, and let $L_{val}(\theta(\mathbf{k}))$ be the corresponding evaluation loss. The outer objective is

$$\mathbf{k}^* = \arg\min_{\mathbf{k} \in \Delta_n} L_{val}(\theta(\mathbf{k})) \tag{3}$$

This objective naturally induces a bilevel structure: the outer decision over $\mathbf{k}$ depends on the inner-trained $\theta(\mathbf{k})$, while the inner optimum itself is determined by the given mixture. Making this explicit,

$$\theta(\mathbf{k}) = \arg\min_{\theta} \mathbb{E}_{x \sim \mathbb{P}_{\mathbf{k}}} \big[ \ell(x; \ \theta) \big], \quad \mathbf{k}^* = \arg\min_{\mathbf{k} \in \Delta_n} L_{val}(\theta(\mathbf{k})) \tag{4}$$

where $\ell(\cdot; \ \cdot)$ is the per-example training loss. For clarity, we write $\theta(\mathbf{k})$ as the minimizer of the population risk under $\mathbb{P}_{\mathbf{k}}$; in practice, LLMs pretraining is carried out under finite compute and data budgets, and $\theta(\mathbf{k})$ is realized by an approximate solution obtained via stochastic optimization (e.g., mini-batch SGD).

Despite the conceptual clarity of Eqs. (3)-(4), directly solving this bilevel optimization problem is often impractical in modern LLMs pretraining. Accordingly, most existing data mixing methods can be viewed as tractable approximations to Eqs. (3)-(4): they simplify either the outer objective, the inner training dynamics, or both so that mixture optimization becomes feasible in practice.

*Static vs. dynamic mixing*. We distinguish between two settings that will recur throughout this survey. *Static mixing* fixes a single $\mathbf{k}$ before training and uses it throughout. *Dynamic mixing* employs a time-varying schedule $\{\mathbf{k}_t\}_{t=1}^{T}$: training is partitioned into stages, and the mixture is updated at each stage, based on real-time signals such as gradients and losses, or other signals, to better match the model's evolving learning needs.

## 3. TAXONOMY

After presenting a clear formulation of the data mixture optimization problem in the previous section, this section turns to a systematic review at the method level. In recent years, many studies have sought improved data mixing schemes under limited compute and data budgets in order to enhance the efficiency of LLMs pretraining. In earlier practice, some efforts adjusted mixtures manually through repeated trial and error [11], which required multiple full or near-full pretraining runs to compare different configurations and is prohibitively expensive in large-scale LLMs settings. In this section, we do not treat such trial-and-error practices as a separate class of methods. Instead, we focus on data mixing methods that, under given budget constraints, exhibit relatively clear design motivations, reusable strategies, or algorithmic frameworks.

This section proposes a finer-grained classification framework and uses it as the organizing thread to provide as comprehensive an overview as possible of the main data mixing methods proposed to date. For each class of methods, we adopt a performance-cost perspective, analyzing their common characteristics and current limitations and briefly summarizing the core ideas of representative work. In addition, we relate the above formulation of data mixture optimization to method-level practice: for each class of methods, we explain how the solution process of the bilevel optimization is approximated or simplified, thereby making data mixture optimization operational under realistic compute and data constraints.

Figure 1 summarizes our taxonomy of existing data mixing methods. At the most macroscopic level, we first divide existing work into two broad categories, static and dynamic, according to whether the method updates the data mixture during training. At the next level, we further refine these two categories based on the mechanisms that drive the methods, thereby obtaining a stratified taxonomy of approaches.

### 3.1 Static

Static mixing specifies a data mixture before training begins and keeps it unchanged throughout the entire training process. Owing to its relatively simple implementation (compared with dynamic methods that need to be continually updated), this family of methods has been dominant in early research and applications. With respect to their driving mechanisms, that is, how this single mixture used throughout training is determined, existing static methods can be roughly divided into two categories: rule-based methods, which rely on directly usable rules or simple algorithms designed from such rules to specify or adjust the mixture; and learning-based methods, which train additional models to automatically learn better mixtures under given budget constraints.

#### 3.1.1 Rule-Based

Rule-based data mixing methods initially relied mainly on relatively simple heuristic rules. Typical practices include the following. When the goal is to improve overall model performance, one up-samples domains that are widely regarded as higher quality, such as encyclopedic corpora like Wikipedia. When the goal is to achieve balance across multiple domains, one up-samples scarce domains while appropriately down-sampling very large domains in order to mitigate excessive skew in the data distribution toward certain sources [5]. **Uniform sampling** and **proportional sampling** are also widely used mixing strategies based on simple rules. The former assigns the same weight to every domain, whereas the latter assigns weights proportional to the data volume of each domain and is one of the common default strategies in large-scale training [33-34]. Under a fixed total budget, proportional sampling can minimize the extent to which any single dataset is reused too many times, that is, the maximum effective number of epochs, thereby keeping the effective usage of different domains relatively balanced. Because these strategies are extremely simple to implement and tend to behave robustly across different setups, uniform sampling and proportional sampling are still regarded as strong baselines and are frequently used as comparisons in experiments on new methods.

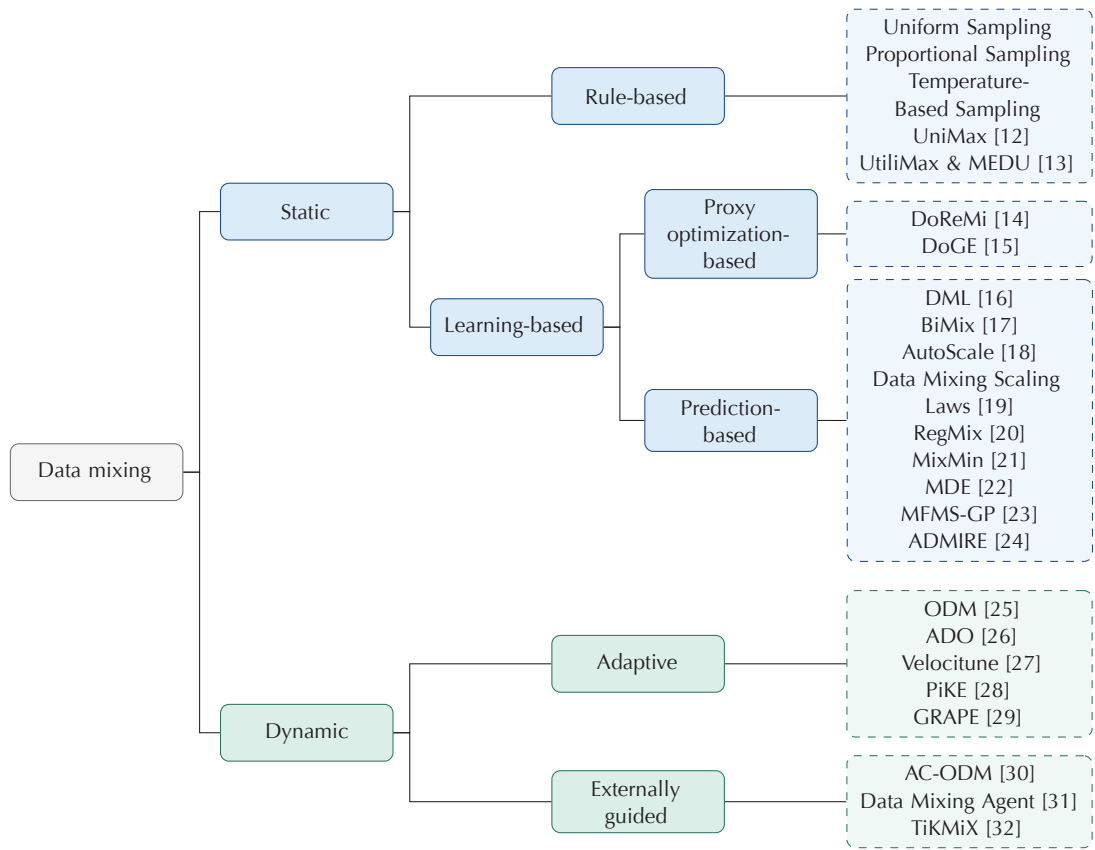

**Figure 1.** Taxonomy of data mixing.

As research has progressed, rule-based methods have also given rise to simple parameterized algorithms while still avoiding additional training, thereby introducing some flexibility between fixed rules and fully learned approaches. A representative example is **temperature-based sampling**, which adjusts the shape of the sampling distribution by introducing a temperature parameter $T$. Given an original domain distribution $p$, temperature-based sampling is usually written as

$$p_i^{(T)} = \frac{p_i^{1/T}}{\sum_j p_j^{1/T}}$$

When $T = 1$, the original distribution is preserved, for example proportional sampling. As $T \to 0$, the distribution becomes sharper and more concentrated on high-probability domains. As $T \to +\infty$, the distribution becomes increasingly flat and eventually approximates uniform sampling. Table 1 provides a summary of a subset of publicly documented LLMs that explicitly report their training sampling temperature $T$, and lists the corresponding $T$ values adopted by each model. Another example is **UniMax** [12], which introduces an upper bound $N$ on the number of repeats. Under the constraint that the number of times each domain is resampled does not exceed $N$, UniMax performs uniform sampling over domains, thereby maintaining balance across domains while effectively suppressing overfitting in low-resource domains. Such parameterized rule-based methods retain the intuitive and interpretable nature of rules, and at the same time endow data mixing with greater tunability and adaptability through a small number of controllable parameters. However, choosing these parameters often becomes a new challenge. For example, it is difficult to find a single temperature $T$ that works reasonably well across different resource scales and domain combinations while not excessively amplifying overfitting risks in low-resource domains. The upper bound $N$ on the number of repeats in UniMax likewise requires careful tuning, since inappropriate settings may degrade performance on high-resource domains.

**Table 1.** Sampling temperature values $T$ reported in the original publications of selected multilingual LLMs.

| Model | $T$ | Model | $T$ |
|---|---|---|---|
| mBERT [35] | 1.43 | VECO [36] | 2.00 |
| XLM-R [37] | 3.33 | XLM-E [38] | 1.43 |
| mBART [39] | 1.43 | Nomic Embed v2 [40] | 3.33 |
| mT5 [41] | 3.33 | mmBERT [42] | $1.43 \to 2.00 \to 3.33$ |

Note: Overall, temperature-based sampling is widely used in multilingual pretraining settings, and the values of $T$ adopted in existing work are to some extent numerically consistent; we therefore compile and present these values with the aim of serving as a reference for subsequent research. In addition to using a fixed $T$ throughout the entire training process, some studies dynamically adjust $T$ as training progresses, so as to more flexibly accommodate the requirements of different training phases. It is worth noting that many studies do not directly use $T$ to denote temperature, but instead adopt $\alpha$ ($\alpha = 1/T$) as the parameter controlling the sampling distribution; in this table, we have consistently converted the $\alpha$ values reported in the original papers into $T$ and listed the corresponding $T$ values.

From the bilevel optimization perspective, early rule-based methods do not attempt to explicitly solve Eq. (3). Instead, they directly prescribe or constrain the mixture $k$ using interpretable heuristic principles, thereby reducing the outer "search over the simplex" to a one-shot rule choice. Furthermore, parameterized rules restrict the decision space to low-dimensional families, so that the mixing strategy can be controlled with only a small number of tunable parameters. Under this view, the bilevel structure is realized in practice primarily by imposing strong constraints on the outer decision space, rather than by repeatedly optimizing through the inner pretraining dynamics.

In general, the core characteristic of rule-based methods is that they can provide reasonably competitive performance while introducing almost no additional cost. Although a large body of recent work has emphasized and empirically demonstrated that the mixtures produced by such methods are often suboptimal, in real-world scenarios where resources are typically highly constrained, these rule-based methods, which are extremely inexpensive, reasonably effective, and have predictable behavior, remain the first choice as starting points and comparison baselines in many research and engineering settings.

It is also worth noting that, as data mixing research has increasingly shifted toward learningbased methods, a new generation of rule-based methods has emerged. Compared with earlier rules that relied only on simple heuristics, these methods employ more complex rule designs and incur higher computational costs, yet they still enjoy a significant cost advantage over fully learning-based methods. Correspondingly, the mixtures they produce are usually closer to optimal and are also clearer in terms of interpretability. For example, **UtiliMax & MEDU** [13] explicitly model data mixing as a portfolio selection problem. UtiliMax, under the Markowitz portfolio framework [43-44], determines the mixture by maximizing the overall utility of domain data while minimizing overall risk, where domain utility scores are estimated via a series of expensive ablation experiments. MEDU (Model Estimated Data Utility) instead uses utility assessments derived from existing LLMs to replace ablation experiments, thereby substantially reducing the cost of estimation. From the bilevel optimization perspective, the above design primarily makes the outer level operational: it represents the outer evaluation of the mixture $k$ using a pair of surrogate objectives—utility and risk—that are more aligned with downstream needs and more interpretable. It further approximates the feedback required by the outer objective via a low-cost, LLMs-based utility estimation pipeline, thereby avoiding repeatedly running expensive inner pretraining for different candidate mixtures. The emergence of these new methods reflects a trend toward pursuing better data mixtures under controllable cost, and to some extent expands the scope of what counts as "rules" in this context.

### 3.1.2 Learning-Based

As discussed earlier, rule-based methods introduce almost no additional cost, but the data mixtures they produce are often only suboptimal. With the development of deep learning, an increasing number of studies have begun to explore learning-based methods of obtaining better mixtures. Such methods typically require training an auxiliary model to learn the data mixture and therefore incur substantially higher cost compared with rule-based methods. However, extensive empirical results show that the resulting mixtures can yield better model performance under a given budget, compared with using rule-based mixtures, especially early mixtures based on simple heuristic rules. Furthermore, some of these methods are also clearer in terms of interpretability.

In practical research and applications, practitioners aim to optimize the data mixture under a limited budget, and it is not feasible to perform full-scale pretraining for all candidate mixtures. For this reason, existing learning-based methods commonly employ proxy models, that is, substitute models whose architectures resemble that of the target main model but whose parameter scales are significantly smaller, and which can to some extent be regarded as reduced versions of the main model. Training around such proxy models to optimize data mixture is essentially an approximate solution strategy under a performance-cost trade-off. While proxy models can reduce training costs, they can only approximately reflect the behavior of the main model under different mixtures and therefore inevitably introduces bias. If the performance loss caused by this bias remains within an acceptable range, one can substantially reduce training cost and, in return, conduct more frequent experiments and pursue more fine-grained algorithmic designs.

More specifically, how proxy models are used in the optimization process is the key to distinguishing different learning-based methods. Existing methods can be roughly divided into two categories: proxy optimization-based methods and prediction-based methods. In what follows, we discuss these two categories in more detail.

*Proxy optimization-based.* Proxy optimization-based methods directly treat the proxy model itself as the optimization carrier. They first optimize the data mixture on the proxy model, and then directly transfer the resulting mixture to the pretraining of the main model. Figure 2 presents an abstract overview of the overall workflow of this class of methods. The implicit assumption behind this class of methods is that the relative ordering of data mixtures is largely preserved between the proxy model and the main model, that is, mixtures that are optimal or near-optimal on the proxy model will also be optimal or near-optimal on the main model. From the bilevel optimization perspective, the key approximation in proxy optimization-based methods is to replace the inner training of the main model with a computationally cheaper proxy training process. Let $\theta_{\text{proxy}}(k)$ denote the parameters obtained by training the proxy model under mixture $k$. With this, $\theta_{\text{proxy}}(k)$ serves as an approximation to $\theta(k)$, so that the response to different mixtures $k$ can be characterized under a controllable cost. Meanwhile, the outer optimization over $k$ is implemented as a stage-wise update procedure: across multiple stages of proxy training, domain weights are iteratively adjusted based on observable signals such as losses and gradients, rather than by precisely solving the full bilevel objective end to end. This avoids repeatedly running expensive full-scale inner pretraining for a large set of candidate mixtures. Representative works include **DoReMi** and **DoGE**, which design different optimization algorithms from the perspectives of robustness and target-domain generalization, respectively.

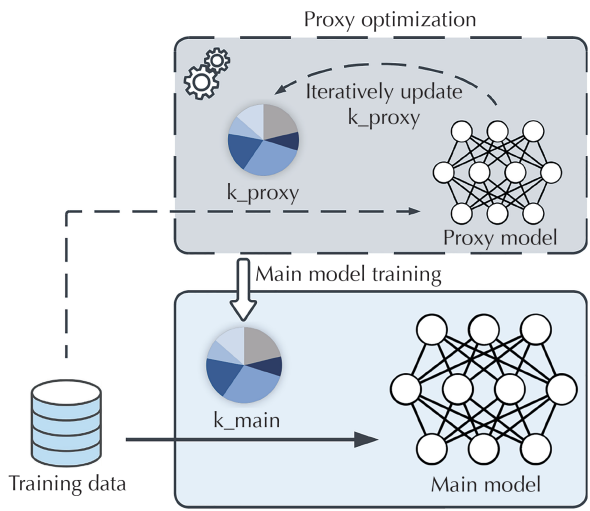

**Figure 2.** Proxy optimization-based methods employ a proxy model as the optimization carrier. An iterative optimization algorithm is executed on the proxy model, which uses internal signals from its training process (e.g., losses and gradients) to drive the optimization and obtain an approximately optimal data mixture. The resulting data mixture is then directly applied to training the target main model.

**DoReMi** [14] (Domain Reweighting with Minimax Optimization) adopts a minimax optimization framework on the proxy model. Its goal is to improve the worst-case performance of the proxy model across all domains, thereby achieving a robustness objective of "not performing poorly on any domain". In terms of implementation, DoReMi trains the proxy model using GroupDRO [45-46] ((Distributionally Robust Optimization). At each update step, it first computes the *excess loss* [47-48] of the proxy model on each domain, defined as the difference between the current proxy model loss and the loss of a reference model that is regarded as performing well; a larger excess loss indicates that the proxy model "learns worse" than the reference model on that domain. DoReMi then uses these excess losses as optimization signals for GroupDRO, assigning larger weights to domains with higher excess loss, so as to strengthen the focus on these domains in the next round of training. After several rounds of iteration, DoReMi averages the domain weights obtained in different rounds to produce a stable domain weight vector, normalizes it, and treats it as a new data mixture that is directly used for pretraining the main model.

**DoGE** [15] (Domain Reweighting with Generalization Estimation) also performs optimization on the proxy model, but shifts the focus to the generalization ability of each training domain to a target domain. Its optimization goal is to maximize the proxy model's performance on the target domain, thereby obtaining a mixture that is more favorable for the target domain. To this end, DoGE designs a gradient-based generalization estimation score for each training domain. It computes the inner product between the loss gradient on the training domain and the loss gradient on the target domain, and uses this to measure their degree of alignment in gradient space. Intuitively, if the inner product is large, then updating model parameters on that training domain tends to move them in a direction that decreases the loss on the target domain, and this training domain exerts a stronger positive transfer effect on the target domain. DoGE therefore assigns higher weights to domains with larger gradient inner products, emphasizing domains that are more "aligned" with the target domain and thereby constructing a target-domain-oriented weighted data mixture.

Overall, proxy optimization-based methods can be regarded as one of the early exploration paths for learning-based data mixing. They demonstrate the practical feasibility of the idea of "obtaining better mixtures through learning" and, according to empirical results, indeed produce mixtures that outperform simple rule-based mixtures, especially early mixtures based on naive heuristic rules. However, this class of methods also suffers from a structural issue that is difficult to ignore. Even when the cost of training the proxy model itself is already nontrivial, achieving a "good proxy" for the main model requires the proxy to be as similar as possible to the main model in terms of model architecture, training configuration, and training data. This means that once these conditions change, for example when the main model architecture is replaced, the training configuration is adjusted, or data domains are added or removed, one has to retrain the proxy model and rerun the entire optimization procedure under the new setting. As a result, transferability is very poor, and the potential for "train once, use many times" is to a large extent restricted; the additional cost is hard to ignore in real large-scale pretraining scenarios. For this reason, subsequent learning-based methods, while inheriting the advantages of this line of work, deliberately weaken their strong dependence on a specific proxy model and instead aim for data mixing learning paradigms with better transferability and reusability, that is, approaches that come as close as possible to "train once, use many times".

*Prediction-based*. To weaken the strong dependence on a specific proxy model and improve the transferability and reusability of learning-based data mixing, prediction-based methods no longer treat the proxy model itself as the direct optimization carrier, but instead use it as a modeling tool. More specifically, these methods assume that, under a given setting, model performance (typically measured by validation loss) has a stable functional relationship with the data mixture used in training. Under a fixed model size and training budget, this can be abstractly written as

$$L_{\text{val}}(\theta(\mathbf{k})) \approx f(\mathbf{k};\ \phi) \tag{5}$$

where $\mathbf{k}$ denotes the data mixture and $f$ is a predictive function parameterized by $\phi$. The goal of prediction-based methods is to learn this functional relationship from a small number of proxy experiments and thereby turn data mixture optimization into an optimization problem over the predictive function. In this process, the proxy model serves to provide a set of "mixture-performance" observations used to fit or train the predictor. Depending on how the function $f$ is modeled, existing work can be broadly divided into two categories: explicit prediction and implicit prediction. Figure 3 presents an abstract overview of the overall workflow of this class of methods.

From the bilevel optimization perspective, Eq. (5) serves as a tractable surrogate for the outer objective by approximating $L_{\text{val}}(\theta(k))$ as a function of $k$. This turns data mixture optimization into a single-level optimization over $k$ on the learned functional relationship or predictor, avoiding the need to explicitly optimize through the expensive inner training process.

**Explicit prediction methods** explicitly model the functional relationship $f$ between model performance and data mixture. They typically specify a parametric functional form with a small number of interpretable parameters and then fit these parameters using observations obtained from training the proxy model under several different mixtures. Representative methods include **DML**, **AutoScale**, **BiMix**, and **Data Mixing Scaling Laws**.

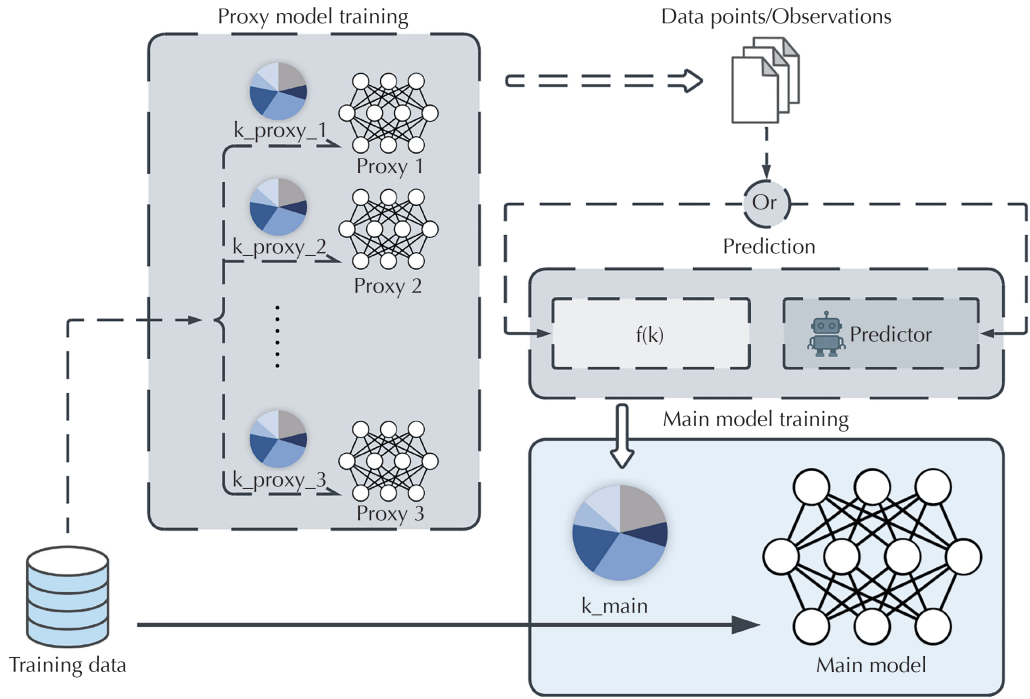

**Figure 3.** Prediction-based methods assume that model performance is a function *f* of the chosen data mixture. Once this function *f* has been learned, one can derive the theoretically optimal data mixture. Such methods are typically instantiated in two ways: explicit prediction and implicit prediction. Explicit prediction methods model *f* as an explicit function and fit it using observations collected from a set of proxy-model experiments. Implicit prediction methods instead treat *f* as a black-box function and learn it by training a prediction model, whose training data consist of the configurations of proxy-model experiments and their corresponding performance metrics.

**DML** [16] (Data Mixing Laws) is one of the earliest explorations in explicit prediction. Under a fixed model size and training step budget, it expresses model performance as a function of the data mixture alone, and uses this function to predict the validation loss associated with different mixtures. Concretely, for each validation task or metric *i*, DML fits a "data mixing law" of the form

$$L_i(\mathbf{k}) = c_i + a_i \exp\left( \sum_{j=1}^{n} t_{ij} k_j \right) \tag{6}$$

where $\mathbf{k} = (k_1, \ldots, k_n)$ is the domain mixture vector, and the parameters $c_i$, $a_i$, $t_{ij}$ are fitted from a set of observed "mixture $\mathbf{k}$ – validation loss" pairs. Here, $c_i$ can be viewed as a mixtureindependent loss floor for metric *i*, $a_i$ scales the mixture-sensitive component, and $t_{ij}$ characterizes how strongly (and in which direction) increasing the proportion of domain *j* affects $L_i$ (e.g., a more negative $t_{ij}$ suggests that upweighting domain *j* tends to reduce the loss on metric *i*). Intuitively, this assumes that the loss curve over mixtures can be approximated by a combination of exponential terms, yielding a function $L(\mathbf{k}) = f_\theta(\mathbf{k})$ at a given scale. Equivalently, the inner sum $\sum_j t_{ij} k_j$ forms a single "mixture score" for metric *i* by linearly aggregating domain proportions, and the exponential maps this score to a smooth loss surface over the mixture simplex, enabling efficient prediction and optimization at a given scale. By nesting this one-variable mixture function within scaling laws [49], DML further extrapolates the loss for different mixtures to larger model sizes and more training steps. This single-variable modeling strategy makes the function and its fitting relatively simple, but it ignores the joint scaling relationship among data mixture, model size, and data volume, which introduces non-negligible error. A simple nesting strategy tends to accumulate these errors and ultimately limits the method's ability to transfer across scales.

**AutoScale** [18] and **BiMix** [17] both recognize that the optimality of a data mixture changes with scale, and therefore do not treat the mixture as the only explanatory variable when modeling the relationship between performance and mixture. Intuitively, they treat the mixture as a "budget allocation rule" and explicitly model how changing the overall budget (tokens or steps) reshapes the loss landscape over the mixture simplex.

In **AutoScale**, under a given total token budget $N$, the authors explicitly assume that the validation loss can be decomposed as a sum of contributions from each domain, and that each contribution follows a power-law form similar to neural scaling laws. Let $k_i$ denote the mixture weight of domain $i$. AutoScale assumes that the component of validation loss associated with domain $i$ satisfies

$$L_i(N, \mathbf{k}) \approx (N_{i,0} + k_i N)^{-\gamma_i} + \ell_i \qquad (7)$$

where $N_{i,0}$ is an "equivalent data volume" constant that captures the indirect contribution from other domains to domain $i$, $\gamma_i$ controls the rate of diminishing returns for that domain, and $\ell_i$ corresponds to the irreducible loss associated with that domain. Here, $k_i N$ can be interpreted as the expected number of tokens allocated to domain $i$ under mixture $\mathbf{k}$ within the total budget $N$, while $N_{i,0}$ acts like a domain-specific "head start" (a pseudo-token count) reflecting cross-domain transfer to domain $i$. Based on this, AutoScale approximates the overall validation loss as

$$L_{\text{val}}(\theta^*(N, \mathbf{k})) \approx \sum_{i=1}^{n} \left[ (N_{i,0} + k_i N)^{-\gamma_i} + \ell_i \right] \qquad (8)$$

where $\gamma_i$ controls the rate of diminishing returns, while $\ell_i$ represents the irreducible loss term.

To learn the parameters $(N_{i,0}, \gamma_i, \ell_i)$ for each domain, AutoScale trains the proxy model under a small number of different mixtures $\mathbf{k}$ and fits the above functional form to the observed validation losses. Once the parameters have been fitted, and for a given target budget $N$, AutoScale directly solves

$$\min_{\mathbf{k} \in \Delta_n} \sum_{i=1}^{n} \left[ (N_{i,0} + k_i N)^{-\gamma_i} + \ell_i \right] \qquad (9)$$

over the simplex to obtain an approximately optimal data mixture under budget $N$. In this sense, the optimization above asks how to split a fixed token budget $N$ across domains (via $\mathbf{k}$) so that the summed predicted per-domain loss curves are minimized. Since the learning of $(N_{i,0}, \gamma_i, \ell_i)$ is decoupled from the specific budget $N$, the same set of parameters can be reused at different budget levels, which in principle supports inferring approximately optimal mixtures across multiple scales.

**BiMix** adopts a more complex modeling strategy and expresses model performance as a bivariate function of data mixture and training steps. Typically, for each domain $i$, BiMix fits a function of the form

$$L_i(k_i, s) = A_i k_i^{\alpha_i}(B_i s^{\beta_i} + C_i) \tag{10}$$

where $k_i$ is the mixture weight of domain $i$, $s$ is the number of training steps, and $A_i$, $\alpha_i$, $B_i$, $\beta_i$, $C_i$ are parameters fitted from multiple observed "$(k_i, s)$ – loss" pairs. Intuitively, $k_i$ is the knob for "how much domain-$i$ data the model sees," while $s$ is the knob for "how far training has progressed"; the exponents $\alpha_i$ and $\beta_i$ quantify how sensitive the loss is to increasing domain-$i$ exposure versus continuing training. With a single bivariate function family, BiMix can predict the loss for different mixtures at different training steps, and jointly characterize how performance evolves with the mixture and with training progress.

**Data Mixing Scaling Laws** [19] advance this line of work by drawing on the idea of the Chinchilla scaling law [49]. They jointly consider data mixture, model size, and data volume as three varying factors and express the loss as a trivariate function $L(N, P, \mathbf{k})$, where $\mathbf{k}$ denotes the mixture, $N$ denotes the training data volume (e.g., the total number of training tokens), and $P$ denotes model scale (e.g., parameter count). For different training budgets, Data Mixing Scaling Laws propose two functional forms. The additive scaling law lets only the irreducible error (bias term) depend on the data mixture—intuitively, the mixture shifts the "loss floor" but does not change the rate at which loss improves as we scale data or model size, and assumes a form such as

$$L_{\text{add}}(N, P, \mathbf{k}) = E(\mathbf{k}) + aN^{-\alpha} + bP^{-\beta} \tag{11}$$

where $E(\mathbf{k})$ captures the irreducible loss under different mixtures, while $a$, $b$, $\alpha$, $\beta$ are independent of the mixture; here $a$ and $b$ set the overall strength of the data/model scaling effects, and $\alpha$ and $\beta$ control how quickly returns diminish as $N$ or $P$ grows. The additive scaling law (i.e., Eq. (11)) thus amounts to assuming that the relative ranking of data mixtures remains essentially unchanged across scales, which leads to a simpler fitting problem with fewer parameters and reduced risk of overfitting. In contrast, Eq. (12) is the joint scaling law: it retains the same interpretation that $a$ and $b$ characterize the overall strength of the data/model scaling effects and that $\alpha$ and $\beta$ govern how quickly returns diminish as $N$ or $P$ increases, but it further allows the scaling terms to depend on the mixture—so different mixtures can change not only the floor $E(\mathbf{k})$ but also "how much you gain" from adding data or parameters, and assumes a form such as

$$L_{\text{joint}}(N, P, \mathbf{k}) = E(\mathbf{k}) + a(\mathbf{k})N^{-\alpha(\mathbf{k})} + b(\mathbf{k})P^{-\beta(\mathbf{k})} \tag{12}$$

meaning that not only the irreducible error term $E(\mathbf{k})$ varies with the mixture, but the coefficients and exponents governing how the loss scales with data volume $N$ and model size $P$ are also determined by the mixture (i.e., the mixture can affect both "where the curve ends up" and "how fast it gets there"). This allows the optimal mixture to depend explicitly on scale. The tradeoff, however, is a more complex functional form and the need for more observations for stable fitting, although in principle it can capture the joint "mixture-scale-performance" relationship more accurately.

**Implicit prediction methods** also assume the existence of a function *f* that maps a configuration to performance, but treat it as a black-box function rather than specifying an explicit analytic form. Instead, they learn this relationship directly through a predictive model. A typical workflow is as follows. First, the proxy model is trained under a series of different configurations, which include different data mixtures and possibly other factors such as model size and training steps. The resulting configurations and their validation performance metrics are collected as data points. Then, a predictor is fitted with configurations as inputs and performance metrics as targets. Finally, this predictor is used to search over the mixture space and approximately solve for the optimal mixture. Representative methods include **RegMix**, **MixMin**, **MDE**, **MFMS-GP**, and **ADMIRE**.

**RegMix** [20] is a representative of methods that formulate data mixture optimization as a regression problem. It directly uses the data mixture as input to a regression model and takes the validation loss observed on the proxy model as the regression target. In most practical implementations of this line of work, LightGBM [50] is used as the predictor. After training, the regression model is used to search over the mixture space and identify a theoretically optimal mixture. The core assumption of RegMix is the *Rank Invariance of Data Mixtures*, that is, the relative ranking of different mixtures on a small-scale proxy model is largely preserved in a larger-scale setting.

**MixMin** [21] and **MDE** [22] (Mixtures of Data Experts) introduce a stronger structural assumption on top of implicit prediction. Under certain idealized conditions, they approximate the predictions of a model trained on mixed data by a linear combination of the predictions of domain-specific models trained on single-domain data, thereby using model mixtures to approximate the effect of data mixtures. Let $\mathcal{D}_i$ denote domain $i$, $f_i$ the domain-specific model trained on $\mathcal{D}_i$, and $f_{\mathbf{k}}$ the model trained on the mixed distribution $\mathbb{P}_{\mathbf{k}} = \sum_i k_i \mathbb{P}_i$. Under the idealized assumption in these methods, for any input $x$, one has

$$f_{\mathbf{k}}(x) \approx \sum_{i=1}^{n} k_i f_i(x) \tag{13}$$

or, in probabilistic form,

$$p_{\mathbf{k}}(y \mid x) \approx \sum_{i=1}^{n} k_i p_i(y \mid x) \tag{14}$$

where $p_i(y \mid x)$ is the predictive distribution learned by the domain-specific model $f_i$ on $\mathcal{D}_i$. This approximation directly captures the core idea of using model mixtures to approximate the result of training on mixed data: instead of retraining a large model under every candidate mixture **k**, one approximates the behavior of these candidate models by linearly combining a small set of expert models {*f*}. At the same time, this formulation trades off fidelity for tractability by assuming that mixed-data training can be well approximated via a linear interpolation of singledomain experts, thereby turning repeated retraining into a directly optimizable objective over **k**.

Under this approximation, the loss of a model trained on mixed data and evaluated on a validation set can be written as

$$L(\mathbf{k}) \approx \mathbb{E}_{(x,\,y)\sim\mathcal{D}_{val}}\left[\ell\left(\sum_{i=1}^{n}k_i f_i(x),\; y\right)\right] \tag{15}$$

which makes the dependence originally implicit in the "train + evaluate" pipeline explicit as a convex objective in **k**. In this way, the data mixture optimization problem can be reformulated as a convex optimization problem. MixMin directly leverages this approximation and applies mirror descent or related convex optimization methods to the loss of these mixed predictions on the validation set, iteratively updating the data mixture. MDE further records the loss of mixed predictions on the validation set together with the corresponding mixtures and uses them to fit a regression model, enriching the regression-based approach exemplified by RegMix with additional structural information and thereby improving predictive accuracy.

**MFMS-GP** [23] (Multi-Fidelity Multi-Scale Gaussian Process) and **ADMIRE** [24] (Accelerated Data Mixture Re-weighting) formulate data mixture optimization as a multi-fidelity, multiscale Bayesian optimization problem and use Gaussian processes [51] to model the relationship between data mixture, scale, and performance. ADMIRE constructs a product kernel composed of an RBF kernel [52] over the mixture space and a downsampling kernel [53] based on model size, and uses it to jointly model the relationship between performance and "data mixture + model size." MFMS-GP goes further by using a product of three RBF kernels to jointly model performance as a function of data mixture, training steps, and model size. Both methods follow the standard Bayesian optimization paradigm: they conduct sequential experiments under a limited budget, continually add new observations, and update the posterior of the Gaussian process, in order to more efficiently approach the optimal data mixture under a fixed budget.

Overall, explicit prediction methods specify a functional form and parameters with semantic meaning, which makes them more intuitive and more interpretable. Implicit prediction methods treat the function as a black box and are more flexible in form. Although they follow different paths, both types of methods weaken the strong dependence on a specific proxy model. They only require a limited number of proxy experiments to learn the relationship between model performance and data mixture, and this learned relationship can then be reused to guide mixture selection. This enhances the transferability and reusability of learning-based data mixing. More importantly, these methods provide an alternative to mixtures set purely by prior beliefs or heuristic intuition. They advocate first characterizing the relationship between performance and mixture, and then optimizing the mixture based on this relationship.

As discussed earlier, the effectiveness of prediction-based methods hinges on how accurately they capture the relationship between model performance and data mixture. In real large-scale pretraining scenarios, however, this relationship is far from being one-dimensional. Existing work has shown joint scaling effects between data mixture and scale, including model size and data volume: the relative optimality of different mixtures is not invariant across scales. A simple example illustrates this: low-entropy domains such as code and mathematics can greatly accelerate learning when the data budget is small, and thus tend to receive higher weights; yet as the training data volume increases, the marginal gains from these low-entropy domains diminish, and more diverse domains such as web data may become more important. In addition to scale, training configuration, per-domain data quality, and mutual promotion or interference among domains also jointly influence performance together with the data mixture. The true relationship between model performance and these variables is highly complex, and it is almost impossible to model it exactly and completely.

Against this backdrop, existing prediction-based methods can still be viewed as products of performance-cost trade-offs. They deliberately ignore part of the influencing factors, sacrificing some accuracy in exchange for more manageable modeling complexity and a feasible fitting or learning process. Even so, the development trajectory of these methods suggests that predictionbased data mixing is generally moving toward models that are more complex but also better able to capture the underlying relationships. This is a trend that future work will need to confront when pushing these methods to larger scales and more complex settings.

### 3.2 Dynamic

Previous studies have shown that the optimality of a data mixture varies with many factors, including training progress. This implies that a single fixed mixture determined before training begins is unlikely to remain optimal throughout the entire pretraining process and cannot adequately accommodate the model's changing needs at different stages. At the same time, for existing learning-based static methods, regardless of how strongly they depend on a particular proxy model, any change in the application setting (for example, modifications to the model architecture, training configuration, or the composition of datasets and domains) will inevitably introduce bias and typically requires relearning the mixture, which leads to non-negligible cost.

Against this background, a natural direction for further optimization is to no longer treat data mixing as a separate stage outside pretraining, but instead integrate it into the pretraining process itself, so that the data mixture can be adjusted dynamically as training proceeds. By partitioning pretraining into several stages and updating the data mixture at each stage according to the model's current state and needs, one can better control the additional overhead within a single optimization run and, at the same time, adapt more easily when the application setting changes, thereby reducing the cost of retuning caused by transfer bias. Along this line of thought, the paradigm of dynamic mixing has emerged, in which the data mixture is allowed to evolve over time.

From the bilevel optimization perspective, dynamic mixing methods can be broadly divided into two families: adaptive and externally guided. Dynamic mixing methods typically do not aim to solve for a single fixed global mixture $k^*$ before training. Instead, they embed the outer optimization over the mixture $k$ into a single continuous pretraining process by partitioning training into multiple stages, maintaining a time-varying mixture schedule $\{k_t\}$, and performing iterative mixture updates at stage boundaries based on feedback from the current model state. In this way, the outer optimization is approximated by a sequence of stage-wise local updates, while the inner training dynamics are incorporated implicitly by continuing training under the current mixture and using observable feedback to inform the next update, thereby avoiding repeatedly running expensive full inner pretraining across many candidate mixtures. The signals that drive these updates are derived from training-time feedback; accordingly, the two families differ in whether the mixture update maps feedback directly (adaptive) or through an explicit controller (externally guided).

In what follows, we discuss these two families in turn. Figure 4 presents an abstract overview of the overall workflow of this class of methods.

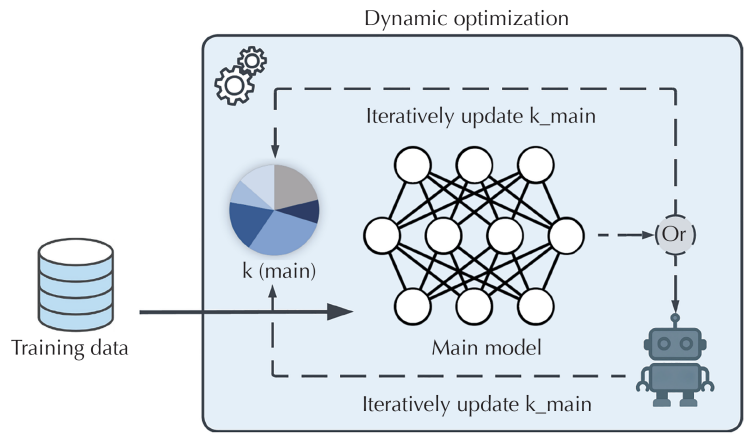

**Figure 4.** Dynamic methods adjust the data mixture on the fly during training of the target main model. Depending on the signal used to drive these updates, they can be divided into two categories. Adaptive methods use internal signals from the target main model (e.g., losses and gradients) as the driving signal, whereas externally guided methods use an external controller to process training-produced signals and output updates.

### 3.2.1 Adaptive

Adaptive methods do not rely on additional external models or offline components. Instead, they update the data mixture solely based on signals generated by the model itself during training, such as training or validation losses, loss gradients, and related statistics. From the bilevel optimization perspective, adaptive dynamic mixing implements the outer-level mixture update by applying a direct update rule that maps training-produced signals to reweighting decisions, thereby replacing the original outer objective with readily observable surrogates. The inner level is realized as standard continued training under the current mixture, while the key approximation is that outer updates are computed from these local signals without differentiating through the full inner training trajectory. Representative methods include **ODM**, **ADO**, **Velocitune**, **PiKE**, and **GRAPE**.

**ODM** [25] (Online Data Mixing) formulates data mixture optimization as a multi-armed bandit problem, where each data domain corresponds to an "arm," and the sampling weight of each domain corresponds to the probability of pulling that arm. To balance exploration and exploitation, ODM adopts a variant of the Exp3 [54] (Exponential-weight algorithm for Exploration and Exploitation) algorithm for online updates. In practice, ODM designs a temporally smoothed reward function based on the training loss to approximately measure the information gain contributed by each domain in the current phase: if a domain has led to larger loss reductions over the recent steps, it receives a higher reward. ODM then updates the domain weights according to these rewards, so that domains with higher information gain are sampled more frequently in subsequent phases, thereby pursuing the objective of maximizing information gain.

**ADO** [26] (Adaptive Data Optimization) introduces scaling laws into adaptive data mixing. Its core idea is that, under a given training configuration, the change in training loss with respect to the number of observed samples in each domain can be approximated by a scaling curve, and the local slope of this curve can be used to measure the benefit of further training on that domain. Concretely, ADO fits a separate scaling law for each domain in a stage-wise manner during training, and in each stage it computes the derivative of the training loss with respect to the number of observed samples, that is, "how much loss reduction a single additional sample can bring." This derivative is treated as the current learning potential of the domain: the larger its value, the higher the marginal gain from continuing to sample that domain. Based on this, ADO allocates more budget in each stage to domains with larger learning potential, thereby improving pretraining efficiency under a constrained overall budget.

**Velocitune** [27] and **GRAPE** [29] (Group Robust Multitarget Adaptive PrEtraining) use learning speed as the core signal. They emphasize that, in a multi-domain setting, one should prioritize domains that are learning the slowest, so as to avoid situations where the model converges too quickly on some domains while remaining persistently underfitted on others. The main difference between the two lies in how they define "speed." In Velocitune, the speed $V$ is defined jointly by the initial loss, the current loss, and a predefined target loss, and in essence measures the learning progress of a domain at the current stage relative to its final goal. Velocitune increases the weights of domains that lag behind in progress, that is, those with relatively low $V$. GRAPE, in contrast, uses the *Rate-of-Improvement (RoI)* to characterize learning speed. It computes RoI by dividing the loss reduction in a stage by the number of training steps in that stage, yielding the loss reduction rate within that stage. When updating the mixture, GRAPE increases the weights of domains with relatively low RoI, thereby dynamically balancing learning progress across domains during training.

Unlike the above methods that construct optimization signals from losses and their derived quantities, **PiKE** [28] (Positive gradient interaction-based K-task weights Estimator) uses gradient information to construct its optimization signal. At each stage, PiKE dynamically adjusts the data mixture based on the norm and variance of the training loss gradients for each domain. The gradient norm is used to measure the "room for improvement" of a domain: a larger norm suggests that further parameter updates on that domain still have substantial potential to reduce loss, so the domain should be given a higher weight. The gradient variance is used to measure the noise level of a domain: a larger variance indicates stronger gradient fluctuations and less stable signals, in which case the weight of that domain should be reduced to avoid spending too much budget on noisy samples. By jointly considering both the strength and stability of gradients, PiKE to some extent balances training gain and training confidence.

Although the representative methods above differ in how they design optimization signals and implement update rules, all the information they use to construct these signals comes from the target model itself during training. There is no need to introduce additional external models, offline experts, or complex proxy systems. This paradigm of relying entirely on internal signals makes adaptive methods notably lightweight and easy to integrate, and the additional computational overhead they incur is typically among the lowest of all existing data mixing methods.

However, the flip side of this lightweight nature is that it imposes strict constraints on the complexity of the optimization mechanism. Because adaptive methods are directly integrated into large-scale pretraining loops, even if the extra computational cost per step or per stage appears negligible, repeated application throughout training can accumulate into a non-trivial overhead. One possible workaround is to use coarser stage partitions, for example updating the mixture only at a small number of key stages, which allows more refined and computationally intensive optimization mechanisms to be used within each stage. Yet such coarse-grained partitioning also reduces the frequency of updates, making it harder for the data mixture to keep pace with rapid changes in the model's needs. Overall, adaptive methods can be viewed as design choices that lie on the low-cost side of the performance-cost trade-off within dynamic mixing. They rely solely on internal signals from the model and update the mixture via relatively simple mechanisms. Although their modeling capacity and optimization granularity are limited, they achieve a relatively lightweight implementation that remains feasible in large-scale pretraining scenarios.

### 3.2.2 Externally Guided

Compared with adaptive mixing, externally guided mixing performs stage-wise updates of the data mixture under the explicit guidance of an external controller model. This controller can be pre-trained and kept fixed during pretraining of the target model, or it can be updated dynamically as training progresses, adjusting its own parameters in response to new information provided by the target model during pretraining. From the bilevel optimization perspective, externally guided dynamic mixing realizes the outer-level mixture update through an explicit external controller that takes training-produced observations as input and outputs reweighting decisions, i.e., the outer update is mediated by a decision module rather than a hand-specified mapping. The inner level remains standard continued training under the controller-proposed mixture, and the main approximation is that the controller optimizes a surrogate outer objective defined on these observable signals (possibly with constraints), again avoiding differentiation through the full inner training trajectory while retaining iterative, training-coupled updates. Representative methods include **AC-ODM**, **Data Mixing Agent**, and **TiKMiX-M**.

Both **AC-ODM** [30] (Actor-Critic based Online Data Mixing) and **Data Mixing Agent** [31] construct the controller using ideas from reinforcement learning (RL) and adopt an actor-critic architecture [55]. In this architecture, the policy network (actor), which selects actions, is separated from the value network (critic), which evaluates states or actions and produces TD error signals, and the two components are trained jointly. Once the actor-critic network has been trained, the actor is treated as a controller for data mixing: during pretraining of the target model, it periodically reads the current state and outputs the data mixture for the next stage. The main differences between the two methods lie in how the actor-critic network is trained and how the reward signal is defined. AC-ODM trains the actor-critic network online on a proxy model, where the reward is designed based on gradient alignment of the loss. In contrast, Data Mixing Agent adopts an offline reinforcement learning paradigm. Concretely, it first collects a large number of "data-mixing update trajectories" on a proxy model, where each trajectory consists of a sequence of state information together with the corresponding downstream evaluation feedback, forming an offline dataset. On this dataset, Data Mixing Agent uses CQL [56] (Conservative Q-Learning) to train the actor-critic network, with reward signals defined directly in terms of downstream evaluation metrics.

The design of **TiKMiX-M** [32] draws on methods based on implicit prediction. This work extends the notion of influence functions by introducing domain-aggregated *group influence*, which measures the marginal effect on validation performance when a given training domain is upsampled. Based on this idea, the authors propose two variants, TiKMiX-D and TiKMiXM. TiKMiX-D directly uses group influence to dynamically update the mixture during training: domains that are more "useful" for validation tasks (that is, with larger group influence) are assigned higher weights in subsequent stages. Since group influence is constructed entirely from internal model signals, originating from training and validation loss gradients together with an approximate inverse of the training objective's Hessian, TiKMiX-D essentially still follows the adaptive paradigm. TiKMiX-M, on the other hand, does not directly use the mixture produced from group influence as the final update. Instead, it treats this mixture as a reference mixture. In each stage, TiKMiX-M samples local perturbations around this reference mixture in the mixture space, runs a short training phase for each perturbed mixture, and records the resulting performance, thereby obtaining a set of mixture-performance pairs. TiKMiX-M then fits a LightGBM predictor on these data and uses this predictor to search the mixture space for an approximately optimal mixture for the next stage, which is then used to guide subsequent updates. Compared with TiKMiX-D, TiKMiX-M uses an external prediction model to guide dynamic updates of the mixture and therefore falls under externally guided mixing.

Unlike adaptive methods, externally guided methods do not integrate all optimization components fully and at fine granularity into the pretraining loop of the target model. Instead, they drive mixture updates through a relatively independent controller model, which to some extent avoids the cumulative overhead that would result from repeatedly executing complex computations at every training step or every stage. This design allows the controller to incorporate richer information sources, such as long-term training behavior or downstream task feedback, and to support more refined algorithms and more complex optimization procedures. Compared with adaptive methods, externally guided methods tend to lie on the high-performance side of the performance-cost trade-off within dynamic mixing: they sacrifice part of the simplicity and cheapness of tight integration in exchange for higher modeling accuracy and, ideally, data-mixing updates that are closer to optimal.

At the same time, this paradigm shifts the central challenge to the choice and construction of the controller itself. Ensuring that the controller remains relatively inexpensive while still providing sufficiently effective guidance for data mixing is an important direction for future improvement. Moreover, experience from existing methods suggests that controllers trained on proxy models often unavoidably inherit a dependence on those proxy models. When the target model and the proxy model differ substantially in architecture, training configuration, or data distribution, this dependence can weaken the controller's ability to generalize to new settings and thereby reduce its effectiveness in guiding dynamic updates of the mixture.

## 4. CHALLENGES AND FUTURE DIRECTIONS

As an important means of improving pretraining efficiency under realistic budget constraints, data mixing has in recent years been driving increasingly efficient LLMs training pipelines. Although research on data mixing has been growing and has achieved a series of substantive advances, several key issues remain insufficiently resolved, which to some extent constrains further development in this area and leaves the trajectory of future progress still rather unclear. To conclude this survey, this section discusses the main limitations and challenges faced by existing methods and, on this basis, combines these limitations with our still limited understanding to offer several preliminary views on possible future research directions, with the aim of providing some reference for subsequent discussion and progress in this area.

### 4.1 Limited Transferability

From current research trends, both static and dynamic learning-based data mixing methods are likely to become the dominant direction in the future. Under given budget constraints, they can usually discover mixtures that are significantly better than simple rule-based strategies, as has been demonstrated in an increasing number of empirical results. However, these methods almost always suffer, to varying degrees, from constraints imposed by various factors, and their transferability is often far from ideal. A mixture that performs very well under one specific setting may no longer be effective when transferred to a different model, configuration, or data environment. This "exhibits strong performance in one setting but fails to generalize to others" phenomenon is one of the main reasons currently hindering the further promotion and deployment of learning-based methods. Moreover, limited transferability is not a problem confined to a single dimension. Rather, it arises from heterogeneity and bias across multiple levels, including optimization objectives, training scenarios and configurations, domain partitioning schemes, and even the datasets and validation sets on which these methods rely. This also makes it very difficult to conduct broad and fine-grained horizontal comparisons of existing methods under a fully unified perspective.

In this survey, we have already attempted, from a macroscopic methodological perspective, to categorize and unify existing work, and there are also preliminary attempts in the literature to build unifying views at other levels. For example, **Aioli** [57], from a functional perspective, unifies part of the existing methods into a Linear Mixing Optimization (LMO) framework. This framework sets the optimization objective as minimizing the overall loss across training time steps and assumes that, in static and dynamic methods, the loss is respectively linear or log-linear in the mixing proportions. Based on this, it introduces a family of parameterized functions of the mixing proportions, referred to in the paper as mixing laws. Under this perspective, the authors further argue that the shortcomings of many existing methods mainly stem from inappropriate choices of the parameters of the mixing laws, and they accordingly propose the new method Aioli, which dynamically fits these parameters during training in order to guide data mixture optimization more accurately. Although the set of specific methods covered by this work is still limited at present, it can be regarded as a valuable first step in the broader trend of revisiting and comparing existing methods within a unified framework.

Overall, to systematically mitigate the problem of limited transferability, it is unlikely to be sufficient to make isolated improvements to only one or two local components. From optimization objectives, model architectures and training configurations, and ways of defining domains, to validation sets and evaluation protocols, constructing a unified framework or paradigm that is as complete as possible and mutually compatible across these key dimensions is likely to be an important prerequisite for making this family of methods more mature and widely applicable. In what follows, we discuss these issues from several perspectives, analyzing in detail which kinds of heterogeneity and bias factors are constraining the transferability of existing methods and, drawing on recent research progress, we further explore corresponding avenues for improvement and future development.

### 4.1.1 Heterogeneity in Domain Definitions

Data mixing is essentially about adjusting the sampling weights of different domains. However, most of the methods discussed above tend to focus only on how to mix, and pay much less attention to another prior question: *what is a domain?* In other words, how are domains actually partitioned and defined? Existing methods usually default to using the data source as the basis for domain partitioning, which is both the earliest and still the most mainstream practice. However, such source-based partitioning is often too coarse-grained and is likely to be suboptimal in many scenarios. **Topic Over Source** [58] provides direct evidence for this point: instead of using source as the criterion for domain partitioning, it constructs domains around topics. Concretely, it first uses a BGE model to embed the data and performs multi-stage K-means clustering in the embedding space; it then leverages an LLMs to generate and consolidate topic labels for each cluster; finally, it fine-tunes a classifier to assign topic labels to the full dataset, thereby obtaining a topic-based domain partition. Experimental results show that, under the same data mixing method and the same training budget, this finer-grained domain partitioning that more closely reflects semantic structure leads to better performance.

Along this line, a number of recent studies have begun to explore the dimension of how to partition or define domains. Overall, the technical approaches they use to optimize data mixture can still be placed within the taxonomy proposed earlier and are essentially similar to existing methods; the key difference is that they no longer rely solely on source-based partitioning, but instead redefine what constitutes a domain. Based on current progress, this line of work can be roughly divided into two paths. One path focuses on implicit structure: it uses embedding representations, gradient information, or label information (including human annotations and labels generated by LLMs) and performs K-means [59-60] or related clustering operations in these feature spaces to obtain new domain partitions [61-67]. Data mixing methods are then directly combined with these new domains to further improve performance, or existing ideas are adapted to design new optimization strategies in this redefined domain space. The other path tends to adopt more fine-grained and interpretable partitioning schemes. For example, **Skill-It** [68] empirically demonstrates the existence of a skill-level granularity and constructs a skill graph based on the degree of positive promotion between different skills, then designs optimization strategies at the level of this skill graph.

To sum up, although the currently dominant source-based domain partitioning is simple and easy to use, it is often overly coarse-grained and introduces non-negligible heterogeneity across different studies, making the domain definition itself an important latent variable that affects transferability. An ideal domain partition should at least satisfy two basic requirements: first, data within a domain should exhibit high consistency in terms of distribution and semantics; second, different domains should differ sufficiently in distribution, style, or task properties so as to provide meaningful dimensions for data mixing. In this sense, proposing finer-grained, more principled, and as far as possible reusable paradigms for domain partitioning is one of the key prerequisites for alleviating heterogeneity in domain definitions, improving the transferability of data mixing methods, and promoting their further development.

### 4.1.2 Heterogeneity in Optimization Objectives

Heterogeneity in optimization objectives is also one of the important factors that limit the transferability of data mixing methods. Different optimization objectives often correspond to very different optimization procedures and outcomes. A direct example is that when the goal is to balance performance across domains or to improve overall robustness, the optimization process naturally tends to produce a mixture that is closer to an average allocation, and may even deliberately tilt toward domains that are harder to learn and require more training resources. In contrast, when the goal is a specific domain or a specific task, the resulting mixture will inevitably concentrate on this target domain, and the attention paid to other domains and tasks will correspondingly decrease.

The training stage in which data mixing is applied also leads to significant differences in optimization objectives. The discussion above has mainly focused on the pretraining stage, which is also the primary application scenario for current data mixing methods. At this stage, the primary goal of the model is to improve generalization as much as possible, so data mixing usually places more emphasis on diversity. This also explains why many methods, while advocating increased attention to a few key domains, nevertheless introduce various "floor" or "smoothing" mechanisms in their algorithmic design, in order to avoid overly sharp sampling distributions and to maintain a basic level of diversity in the training data. When the application scenario is no longer limited to pretraining, however, the optimization objective changes, and the design considerations for data mixing change as well. For example, in continual pretraining, avoiding catastrophic forgetting becomes one of the core objectives, whereas in the fine-tuning stage the optimization is often organized around only one or a small number of target tasks.

Recently, some studies have begun to explore data mixing designs beyond the pretraining stage. For example, **IDEAL** [69] applies a scaling factor to the original sampling weight of each domain and uses influence functions to estimate the marginal effect on overall performance of slightly changing the sampling proportion of each domain. Based on these estimates, it dynamically updates the scaling factors, and in this way implements an adaptive dynamic mixing method applied in the SFT stage. Another example is **AMA** [70] (AutoMixAlign), which trains an expert model for each domain and then uses these experts to guide dynamic updates of data mixing during training of the main model. Concretely, it adopts a minimax optimization perspective and uses the main model's excess loss on each domain as the optimization signal. AMA is thus an externally guided dynamic mixing method applied in the DPO stage.

If we hope that changes in optimization objectives do not force us to substantially modify, or even to redesign and completely rerun, the entire data mixing procedure every time, then reducing the strong dependence on any single specific optimization objective becomes a key issue for improving transferability. One possible direction is to design mechanisms that require only minor adjustments when the optimization objective changes and that can even adapt to different objectives without modifying the method itself or incurring substantial additional computation. **Chameleon** [71] provides a useful example. It computes KRLS [72] (kernel ridge leverage scores) for the domain embeddings obtained from a proxy model, and uses these scores to measure the "representativeness" and "uniqueness" of different domains. When the optimization objective is to improve general-purpose performance, Chameleon selects more "representative" domains (those with higher inverse KRLS), whereas when the objective is to improve performance on a specific target task, it places greater emphasis on more "unique" domains (those with higher KRLS).

### 4.1.3 Heterogeneity in Model and Configurations

As discussed earlier, heterogeneity in model architectures and training configurations likewise constrains the transferability of data mixing methods, and this effect is particularly pronounced for methods that rely on proxy models. Although the overall trend in recent work is to gradually reduce the strong dependence on specific proxy models, for example through prediction-based static methods and externally guided dynamic methods, as long as the optimization process is still organized around one or more pre-constructed proxy models, it will inevitably suffer varying degrees of performance degradation when the model architecture or training configuration changes. By contrast, among existing methods, the ones that do not require any additional proxy model and instead update the data mixture directly on the target model using online signals are mainly adaptive dynamic methods, whose dependence on specific architectures and training configurations is relatively weaker in form.

On the other hand, model architectures and training configurations themselves are highly mutable. From base model size and structure to optimizer choice, learning rate schedules, and regularization strategies, almost every component may change frequently in response to hardware conditions, application scenarios, or engineering requirements. For this reason, building a fixed and unified paradigm that is tied to particular architectures and configurations is not realistic in practice. Against this background, if we wish to alleviate the transferability bottlenecks caused by heterogeneity in models and configurations, a more realistic direction is to push data mixing methods further toward adaptive dynamic paradigms, that is, to reduce reliance on additional proxies as much as possible and to base the decision logic more directly on training signals that are observable from the target model itself. In fact, looking at new work over roughly the past year, a considerable number of methods have already begun to adopt this kind of adaptive dynamic design, which to some extent corroborates this developmental trend.

### 4.1.4 Biased and Unrepresentative Validation Sets

Finally, the choice of validation set is likewise an important factor that limits the transferability of data mixing methods. Since most existing methods use validation loss as the optimization signal to adjust the data mixture, this naturally introduces a critical dependency: the validation set itself must be sufficiently representative and should reflect the data distribution in the target application scenario as faithfully as possible. Even if domain partitioning and method design are reasonably well done, if the validation set exhibits systematic bias or covers only a limited range of scenarios or task types, the data mixture optimization it drives is likely to be misled, and the resulting mixture will be difficult to reuse in other settings.

From the perspective of validation set construction, future work on constructing and selecting validation sets should place greater emphasis on their ability to cover the diverse problem scenarios that may arise in practice, rather than evaluating only a small number of typical cases. Only by ensuring, at the level of the validation set, as thorough a characterization as possible of the target distribution and its variants can we avoid the original validation set becoming completely invalid or severely distorted when the problem scenario changes, thereby providing a more reliable safeguard for data mixing methods to maintain effectiveness and transferability across different settings.

### 4.2 Unstandardized Evaluation Protocols and Benchmarks

As discussed above, the transferability of existing data mixing methods across settings remains constrained by multiple factors. At the same time, differences in evaluation protocols and benchmarks across studies also reduce cross-paper comparability in practice, which in turn limits unified and rigorous quantitative analyses of existing methods. This subsection focuses on several key evaluation-related dimensions. We summarize common practices and their variations in the literature, and provide a few preliminary discussions, with the goal of offering a reference point for more systematic standardization efforts in future work.

### 4.2.1 Heterogeneity in Optimization and Update Signals

In data mixing research, evaluation quantities are used not only to report final outcomes, but are also often used directly as signals to drive mixture optimization or updates. Different studies vary substantially in how such signals are chosen and constructed. At a coarse level, existing signals can be grouped into two categories.

The first category consists of loss-based signals. These methods typically use loss on a validation or development set as the direct signal, or derive indirect signals from the loss, such as excess loss [14, 70], the rate of loss change [29], or the expected loss decrease [26]. Some work further constructs signals using first-order or second-order information related to the loss, such as gradient inner products or variances [15, 28], and influence estimation [32, 69]. Because these signals are tightly coupled to training dynamics and to how the validation set is constructed, their comparability often relies on consistent evaluation-set design and a consistent definition of the budget.

The second category consists of downstream-metric-based signals. These methods use downstream task metrics, such as accuracy or few-shot scores, as direct signals, or construct more abstract proxies based on them, such as data utility [13]. Compared with loss-based signals, downstream metrics are more closely aligned with application objectives. However, their stability and reproducibility are often more sensitive to the choice of task sets, prompting and evaluation details, and statistical noise.

### 4.2.2 Heterogeneity in Distributional Settings

From the perspective of the relationship between evaluation and training distributions, common evaluation settings in existing work can be roughly summarized into three types.

The first type is matched distributions, where the evaluation distribution is consistent with, or highly similar to, the training distribution. This is usually implemented via the same domain partitioning scheme and similar domain-weight specifications. Under this setting, evaluation results more directly reflect how well a method optimizes the intended training objective, but they can also be more affected by particular choices of domain partitioning or data processing procedures.

The second type is subset coverage, where evaluation covers only a subset of the trainingdomain set. For example, training mixtures may include multiple domains, while evaluation focuses on only some domains or capability dimensions. This setting can highlight gains for specific targets, but cross-paper comparisons require a clear statement of how evaluation domains are selected and how they correspond to the training mixture.

The third type is out-of-distribution evaluation, where evaluation domains do not overlap with training domains, or where they appear similar at the surface level but differ substantially in corpus sources, task formats, language, or style. This setting is closer to testing generalization or transfer. However, because different studies define and construct "domains" in different ways, OOD results typically require careful interpretation when used for horizontal comparisons.

### 4.2.3 Heterogeneity in Cost Accounting

Following the performance-cost perspective adopted throughout this paper, evaluation should consider not only model performance, but also the additional cost introduced by a method. Yet existing work does not follow a unified practice in how cost is measured and reported. Some studies use the number of training tokens or training steps as the primary budget definition, while others prefer FLOPs or wall-clock time. These measures are not strictly interchangeable, since their relationships can depend on sequence length, batch size, parallelization strategies, and implementation details. As a result, stating the cost definition explicitly, and aligning cost accounting as much as possible, is practically important for cross-study comparisons.

Moreover, cost evaluation is often coupled with how a method is integrated into the pretraining pipeline. For some methods, the main cost is incurred in an offline one-time stage, which may be reusable under certain conditions. For other methods, changes in the setting may require rerunning all or part of the key procedures. Therefore, in addition to reporting the cost of a single run, specifying which components must be rerun and which components can be reused under setting shifts can help characterize practical usage cost more accurately.

In general, unified and rigorous evaluation protocols and benchmarks remain an open direction for the community to explore. The summary above should be viewed as an initial framework. Future work can build on it by more precisely defining evaluation goals and distributional settings, clarifying budget definitions and reporting items, and conducting more systematic horizontal comparisons under controlled variables. Such efforts would gradually improve comparability across data mixing methods and the reproducibility of empirical conclusions.

### 4.3 Trade-off between Performance and Cost

As repeatedly emphasized earlier, given real-world resource constraints, almost all existing data mixing methods are in practice compromises shaped by a performance–cost trade-off. This also constitutes another key factor limiting further progress in this area: at the current stage, it is difficult to identify a data mixing scheme that is both approximately optimal and sufficiently inexpensive. Methods that achieve better performance tend to be very costly, whereas cheaper methods usually fall short in terms of effectiveness.

The reasons for this situation are manifold, but two aspects are particularly salient. In what follows, we discuss these two aspects in turn and, on this basis, offer some preliminary views on possible directions for improvement.

#### 4.3.1 Inherent Limits of Learning-Based Methods

From an applied perspective, the learning models on which learning-based data mixing methods rely have limited capacity, and this constitutes a key constraint in the current performance-cost trade-off. Whether they serve as proxy models that act as optimization carriers, as predictors, or as external controllers that guide dynamic mixing, the models used in existing methods can, under a limited budget, typically provide only a coarse approximation to the true relationship between data mixtures and model performance, and they find it difficult to achieve both accuracy and stability while keeping costs under control.

Existing work has already explored a variety of model types. Proxy optimization-based methods typically adopt a smaller-scale version of the target main model as the optimization carrier, that is, a model with the same Transformer architecture, so as to remain structurally close to the main model. By contrast, the predictors used in prediction-based methods and the controllers that guide updates in externally guided dynamic methods are usually much simpler models, such as multilayer perceptrons, LightGBM, or Gaussian processes based on kernel functions. Nevertheless, there is still no consensus on a question that appears basic but is in fact crucial: in the specific setting of data mixing, which class of models is most suitable as the learner, in the sense of capturing the relationship between data mixtures and target performance as accurately as possible while incurring only modest computational and data costs?

Identifying such an optimal learner clearly remains an open problem. One point that seems relatively clear is that Transformer-based models, due to their architectural complexity and high training cost, are unlikely to be a universally preferred option under strict budget constraints; by contrast, simpler and more lightweight model architectures appear much more realistic in this setting. Future progress is therefore likely to depend on two directions: first, identifying within existing families of lightweight models those architectures and designs that are particularly well suited to this task; and second, developing new architectures with inductive biases tailored to the data mixing problem, potentially including specialized learning models designed specifically for this purpose, in order to achieve a more reasonable balance between expressive power and resource consumption.

### 4.3.2 Underdeveloped Theoretical Foundations

From a theoretical perspective, underdeveloped foundations are another major factor limiting the further progress of data mixing methods when viewed from the perspective of the performance–cost trade-off, because how a method is designed, and how far it can in principle be pushed, depends heavily on the underlying theoretical framework it can rely on. Many existing methods rely on assumptions that are often restrictive. These assumptions often arise from approximate theoretical derivations or idealized settings and are difficult to guarantee as both reliable and stable across different configurations in realistic large-scale pretraining scenarios.

Other methods appeal to more intuitive and seemingly more robust theoretical considerations, such as the prior or empirical belief that assigning a larger sampling weight to a domain will usually lead to better performance on that domain, or the scaling laws [49, 73] that have been widely discussed in recent years. Scaling laws suggest that, within a certain range, model loss often exhibits an approximate power-law relationship with model size, training data volume, and training compute, and can thus provide a reference for assessing the expected gains from scaling. However, such results are still largely qualitative or based on coarse-grained curve fitting, and they offer limited precise quantitative guidance on how to allocate data and compute across domains under a fixed budget. Moreover, there is still a lack of theoretical work that is specifically tailored to the data mixing problem itself and that has been widely accepted as a unified foundation.

Taken together, these gaps at the theoretical level converge to a core issue: we lack a sufficiently clear and broadly accepted theoretical basis to answer three basic questions, namely, what kinds of data mixtures should be regarded as good, why they are good, and how such mixtures should be systematically derived under a given budget constraint. Correspondingly, we also lack a unified theoretical framework that characterizes, within the performance-cost trade-off, the structure and properties that an optimal data mixing method ought to have.

To overcome this limitation, it is clearly not sufficient to work only at the application level, for example by replacing or refining the learning models used. Progress on the theoretical side is equally indispensable. Future theoretical research will need, on the one hand, to move beyond existing empirical observations and scaling laws toward more precise quantitative analyses, and, on the other hand, to develop theories and analytical paradigms that are explicitly targeted at the data mixing problem itself, rather than simply transplanting results obtained in other settings. Only with substantive advances at this level can we provide stronger theoretical support for the design of subsequent methods and, in turn, move closer to genuinely approximately optimal solutions under performance and cost trade-offs.

### 4.4 Exploratory and Underexplored Directions

Finally, we discuss several directions that are exploratory in nature and have not yet been studied in depth. First, these include research pathways for which preliminary attempts already exist but which have not yet led to a mainstream consensus. Second, they include possible future trajectories that we propose based on our analysis of existing work. We do not claim that these directions necessarily represent the correct path or ultimate destination for research on data mixing. Rather, by organizing and presenting these possibilities, we hope to offer complementary perspectives and, beyond the methods surveyed so far, further stimulate new questions and lines of discussion.

#### 4.4.1 Finer-Grained Data Mixing

Previously, when discussing the question of *what is a domain*, we pointed out that finergrained domain partitions can often significantly improve the performance of existing data mixing methods without changing the methods themselves. This observation naturally raises a further question: if data mixing is designed to act directly on finer-grained objects, can its effectiveness be improved even more? A straightforward extension is to move from the domain level down to the sample level and implement finer-grained data mixing by reweighting individual data samples. Indeed, some existing studies have already explored such sample-level mixing schemes [74-76].

However, this move toward finer granularity brings not only potential benefits but also clear costs. As emphasized in the introduction, from a data-centric perspective there are roughly two main routes to improving pretraining efficiency: the first is data selection at the sample level, which improves dataset quality by filtering individual samples; the second is data mixing at the domain level, which improves overall utilization efficiency by adjusting sampling weights across domains. The former operates at a finer granularity and is in principle more flexible, but is usually more expensive in large-scale corpus settings; the latter is coarser-grained but easier to apply to large-scale pretraining. Against this background, if data mixing is pushed all the way down to an excessively fine level, it is likely to lose the very advantage that makes it suitable for large-scale scenarios in the first place, namely its relatively favorable computational and engineering profile.

From a longer-term perspective, the trend toward finer granularity remains an important source of inspiration. A more realistic direction may be to seek intermediate granularities between samples and domains, so as to strike a better balance between performance gains and additional cost. On the one hand, as discussed earlier, data selection and data mixing are often used jointly in practical applications, and a finer-grained data mixing mechanism would naturally interface better with sample-level data selection strategies. On the other hand, data management and scheduling strategies at different granularities may in the future be designed as a hierarchical, nested process: more sensitive filtering or reweighting is carried out at finer granularities, while more stable budget allocation is performed at coarser granularities. In such a multi-level strategy, finer-grained data mixing would form a key link that connects sample-level decisions with domain-level decisions and is therefore a direction that merits more systematic exploration.

### 4.4.2 Inverse Design of Data Mixtures

Compared with the forward-designed data mixing methods discussed earlier, recent years have seen the emergence of a relatively uncommon but conceptually suggestive line of work that seeks to understand data mixtures from an inverse perspective. Existing studies typically assume that the data mixture is known, and then ask how to design a better mixing strategy. The inverse perspective instead asks from the opposite direction: for high-profile models that are widely regarded as strong but whose data configurations are fully closed, can we, without direct access to their training data, approximately infer what data mixture they used? In other words, the goal is not to decide how to set the mixture, but to infer the mixture from the model itself or from components associated with it.

**Data Mixture Inference Attack** [77] is a representative example of this line of work. It treats the merge list of a BPE tokenizer [78] as a kind of fingerprint of the training corpus distribution, on the assumption that each pair of tokens that is merged at a given step reflects, to some extent, its relative frequency in the mixed corpus. Concretely, the method replays BPE training separately on each candidate domain in order to estimate the frequency of each token pair within that domain. It then converts the event that a given pair is selected for merging at a particular step into a set of constraints on the domain mixing weights, and finally solves a linear program to identify the domain proportions that best explain the observed merge sequence. In this way, given only a trained tokenizer and a small number of samples from each candidate domain, one can approximately reconstruct the data mixture used at the tokenizer training stage.

Similarly, **Data Proportion Detection** [79] seeks to infer the mixture from observed outcomes. This method uses both the generation distribution of an LLMs and the mixture-performance relationship provided by Data Mixing Laws (DML) to recover the proportional structure of the training data. On the one hand, it posits an exponential relationship between each domain's mixture proportion in the generation distribution and its expected loss. On the other hand, DML provides a functional form for the expected loss of each domain as a function of the training mixture weights. Combining these two components allows the training mixture weights to be expressed explicitly as a function of the mixture proportions in generation. Solving the resulting linear system then yields an approximate reconstruction of the training mixture over domains, under the assumed conditions.

Although at the current stage such inverse methods still face significant limitations in terms of both accuracy and practical usability, they offer a markedly different perspective on data mixing. First, they may provide an indirect way to examine the data composition of closed-source models, helping researchers better understand the rough profile of frontier models in terms of domain coverage, language distribution, and potential biases, and thereby providing auxiliary information for evaluation, fairness analysis, and safety assessment. Second, once it becomes possible to reliably recover certain data mixture patterns that have been empirically validated as effective within a tolerable error range, these patterns themselves can serve as design hints and be used as priors or initialization points for forward data mixture optimization, rather than starting from scratch. Finally, in the longer term, inverse inference and forward design are likely to become complementary: the former helps answer which mixtures appear to work well in practice, while the latter focuses on how to systematically search for better mixtures under given constraints. The combination of the two may ultimately lead to a more closed-loop understanding of the data mixing problem.

### 4.4.3 Pipeline-Aware Data Mixing

Although current data mixing methods have already been applied across multiple training stages, including pretraining, continual pretraining, SFT, and DPO, the mixing strategies used at these stages are largely fragmented and mutually independent. In other words, most existing work focuses on stage-aware local optimization of individual training phases and pays relatively little attention to coordinating the design and evolution of data mixing from the perspective of the entire training pipeline.

Compared with this local, single-stage perspective, we argue that data mixing methods with longer-term planning that explicitly target the full training pipeline should become an important direction for future research. Under such a pipeline-level view, different training stages are no longer regarded as isolated segments, but as a sequential and mutually interacting process. Consequently, the design of data mixing should not only be optimized for the local needs of a single stage, but should also take into account its context. For example, when designing a mixing strategy for the current stage, one should not only satisfy the performance objectives of that stage, but also consider how to create more favorable initial conditions and constraints for subsequent stages.

In this sense, genuinely pipeline-aware data mixing methods have the potential to gradually incorporate the currently scattered, stage-specific mixing strategies into a relatively unified framework. Within such a framework, the mixtures used at different stages would no longer be a loosely concatenated collection of local schemes, but coordinated submodules planned around the overall training objective. How to construct and formalize this pipeline-level perspective is likely to become a key step toward further integrating existing and future methods and systematically improving the efficiency of the overall training pipeline.

## 5. CONCLUSION

This survey examines data mixing for LLMs pretraining from a unified perspective, spanning problem formulation, method taxonomy, and open challenges. Methodologically, we show how prior work can be organized along two main axes—static versus dynamic mixing, and rulebased versus learning-based methods—and further categorize learning-based methods into proxy optimization-based, prediction-based, adaptive, and externally guided approaches. This organization not only provides a systematic view of existing studies, but also highlights the intrinsic connections among design choices in terms of objectives, driving signals, and how they are integrated into the pretraining pipeline.

Taking a step back, several broader insights emerge. First, data mixing has evolved from simple heuristics toward more structured and learning-driven paradigms, yet many methods remain tightly coupled to specific models, datasets, scales, or tasks, which limits their transferability. Second, performance and cost are inherently intertwined: learning better mixtures almost inevitably introduces additional computation, and existing methods are often best viewed as different operating points along a performance-cost trade-off curve rather than definitive, onesize-fits-all solutions. Third, our analysis suggests that future progress will likely require moving beyond fragmented training stages and proxy-specific, signal-driven designs toward more pipeline-aware and theory-guided frameworks, together with more standardized evaluation protocols and benchmarks, enabling tighter coupling between mixture optimization and alignment, safety, and downstream deployment scenarios.

We hope that the formulation, taxonomy, and analysis presented here can serve as a shared reference frame for subsequent research: a foundation for more systematic empirical comparisons, a checklist for designing new methods, and a starting point for developing general principles of data mixture optimization in large-scale LLMs pretraining.

## 6. LIMITATION

Prior studies on data mixing exhibit substantial heterogeneity along several key dimensions, including datasets and preprocessing pipelines, domain partitioning strategies, model architectures and scales, training configurations, optimization objectives, and evaluation protocols and benchmarks. Under such uncontrolled variation, directly juxtaposing the performance or cost numbers reported across papers is often ill-posed and can be misleading: apparent differences may conflate genuine methodological effects with choices in experimental setup, implementation details, and even hardware and distributed-training infrastructure, yielding conclusions that appear precise but are, in fact, not comparable.

For these reasons, this survey deliberately avoids providing leaderboard-style quantitative comparisons or definitive rankings across studies. Our goal is not to produce a single "best" mixture recipe, but to offer a more robust interpretive and analytical framework. Specifically, by presenting a unified problem formulation and a structured taxonomy, we place disparate lines of work within a common coordinate system, and use the performance-cost trade-off as a consistent lens for discussion. We hope this framing helps readers more credibly understand, relate, and select methodological directions under varying objectives and resource constraints.

Looking ahead, more rigorous unified comparisons may become feasible. However, this would require the community to gradually converge on more consistent conventions regarding domain partitioning, data processing practices, training configurations, and evaluation protocols and benchmarks, together with reproducible and shareable reference points. Achieving this, in turn, likely depends on further theoretical and methodological maturation of the field and sustained effort from future work.

## AUTHOR CONTRIBUTIONS

We declare the contributions of each author as follows:

Zhuo Chen:

- Conceptualization: Conceived and defined the survey topic on data mixing for LLMs pretraining and set the overall scope.
- Methodology: Developed the unifying performance-cost analysis perspective and designed the taxonomy that structures the survey.
- Investigation: Performed the literature search and review, and synthesized the research landscape.
- Writing Original Draft: Drafted the initial manuscript.

Yuxuan Miao:

- Visualization: Designed and produced the figures, including the taxonomy overview and the abstract workflow diagrams for major categories of data mixing methods.

Supryadi:

- Conceptualization: Provided substantive input on topic framing and scope, and suggested directions for extending and enriching the discussion.

Deyi Xiong:

- Supervision: Supervised the research and ensured overall quality control.
- Writing Review & Editing: Reviewed the draft and provided guidance for revision, editing, and polishing.

## ACKNOWLEDGEMENTS

The present research was supported by the National Key Research and Development Program of China (Grant No. 2023YFE0116400). We would like to thank the anonymous reviewers for their insightful comments.